\documentclass[conference]{IEEEtran}
\IEEEoverridecommandlockouts
\usepackage{cite}
\usepackage{amsmath,amssymb,amsfonts}
\usepackage{algorithmic}
\usepackage{graphicx}
\usepackage{textcomp}
\usepackage{xcolor}
\usepackage{bm}
\def\BibTeX{{\rm B\kern-.05em{\sc i\kern-.025em b}\kern-.08em
    T\kern-.1667em\lower.7ex\hbox{E}\kern-.125emX}}
\begin{document}

\title{Rethinking Evaluation Metric for Probability Estimation Models Using Esports Data}

\author{\IEEEauthorblockN{Euihyeon Choi}
\IEEEauthorblockA{
\textit{PS Analytics}\\
Seoul, South Korea \\
ceh9856@snu.ac.kr}
\and
\IEEEauthorblockN{Jooyoung Kim}
\IEEEauthorblockA{
\textit{SK Holdings C\&C}\\
Seongnam, South Korea \\
harrykim@sk.com}
\and
\IEEEauthorblockN{Wonkyung Lee}
\IEEEauthorblockA{
\textit{PS Analytics}\\
Seoul, South Korea \\
wonkyung.lee@ps-analytics.com}
}

\maketitle

\begin{abstract}
Probability estimation models play an important role in various fields, such as weather forecasting, recommendation systems, and sports analysis.
Among several models estimating probabilities, it is difficult to evaluate which model gives reliable probabilities since the ground-truth probabilities are not available.
The win probability estimation model for esports, which calculates the win probability under a certain game state, is also one of the fields being actively studied in probability estimation.
However, most of the previous works evaluated their models using accuracy, a metric that only can measure the performance of discrimination.
In this work, we firstly investigate the Brier score and the Expected Calibration Error ($ECE$) as a replacement of accuracy used as a performance evaluation metric for win probability estimation models in esports field.
Based on the analysis, we propose a novel metric called Balance score which is a simple yet effective metric in terms of six good properties that probability estimation metric should have.
Under the general condition, we also found that the Balance score can be an effective approximation of the true expected calibration error which has been imperfectly approximated by $ECE$ using the binning technique.
Extensive evaluations using simulation studies and real game snapshot data demonstrate the promising potential to adopt the proposed metric not only for the win probability estimation model for esports but also for evaluating general probability estimation models.
\end{abstract}


\section{Introduction}
Probability estimation problem is already an important issue in various fields, such as weather forecasting~\cite{https://doi.org/10.48550/arxiv.1912.12132}, recommendation systems~\cite{kweon2022obtaining}, and sports analysis~\cite{lock2014using,robberechts2021bayesian}. 
Since reliable probability in such fields brings great benefits to our life~\cite{zadrozny2001obtaining}, various models have been proposed to estimate the reliable probability.
However, unlike classification problems with true labels, the probability estimation problem is more difficult to evaluate because there is no ground-truth $p$ label~\cite{https://doi.org/10.48550/arxiv.2111.10734}.
It is thus increasingly important to select a metric carefully to measure the performance of probability estimation models adequately. 
As an option of the probability estimation evaluation metric, several works have been done to develop metrics such as the Brier score~\cite{brier1950verification} and the Expected Calibration Error ($ECE$)~\cite{naeini2015obtaining} which are proposed to measure the calibration performance of models instead of accuracy.

Recently with the rapid market growth of esports, research to apply the win probability estimation model to the esports field is also being actively conducted~\cite{hodge2019win,yang2022interpretable}. Among the esports genres, the multiplayer
online battle arena (MOBA) genre is one of the main targets to apply the win probability estimation model based on its high market share in the esports field and also easily accessible data to train the models~\cite{yang2022interpretable}. 
To the best of our knowledge, however, most of the previous works on the MOBA genre are still using accuracy as a metric in the evaluation of their probability estimation models.
Unfortunately in the esports field, using accuracy as an evaluation metric can be more unstable since the datasets in esports have a huge diversity~\cite{hodge2019win} in terms of \emph{operating conditions} (the distribution of ground-truth $p$) which the probability estimation model works with.
In classic sports, the aspect of the game does not differ significantly depending on the point of view of the game.
However in MOBA, the aspect of the game is very different at each point of the game, as the stats of the characters played by each human player change dynamically within the game. This makes the operating condition of datasets completely distinct at each time point.
In addition, repeated updates by game companies make another diversity to the operating condition of datasets by changing the dynamics of the game. 
It is concluded that research on the win probability estimation model using esports data are being conducted on very different datasets without a fixed dataset.
These circumstances make that the suggested model cannot guarantee performance in other cases.

In this work, we provide a detailed analysis of three candidate metrics that can be employed to evaluate probability estimation models for esports data. 
Based on the analysis of candidate metrics, we propose a simple yet effective evaluation metric called Balance score that can address the shortcomings identified in other metrics.
Extensive evaluation using simulation studies and real game snapshot data verifies the benefits of the proposed metric and opens up possibilities to be utilized in general probability estimation model.

\begin{figure*}[ht]
\centering
	\includegraphics[width=\textwidth]{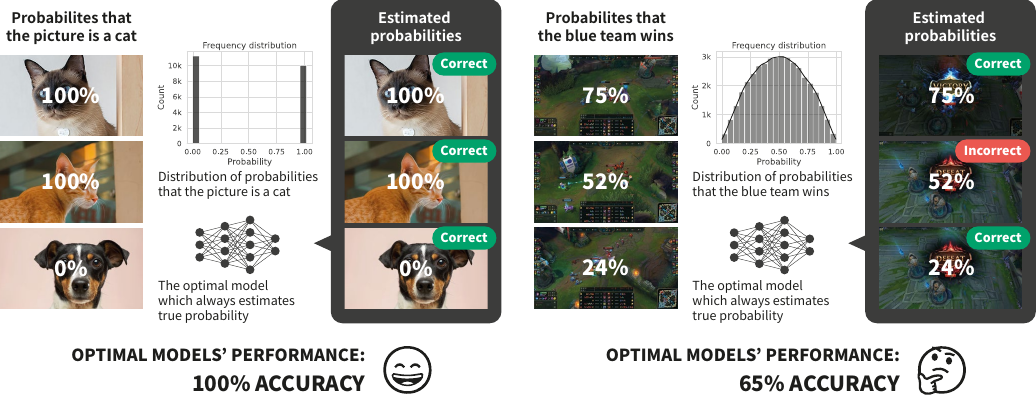}
	\caption{Problem of using accuracy as a measure of probability estimation models in esports. Compare to the image classification task which only ask for the label, even the optimal model may not be able to estimate the result of the match due to the uncertainty of the game snapshot data itself.} 
	\label{figure1}
\end{figure*}

\section{Related Works \& Proposed Metric}

\subsection{Overview}
Most previous studies on win probability estimation models in esports use accuracy as an evaluation metric~\cite{hodge2019win,yang2022interpretable}. 
However, accuracy is a metric that only measures the discrimination performance of the model, so it does not guarantee the performance of the model on estimating the exact win probability~\cite{naeini2015obtaining}.
Also, in general classification tasks such as image classification, training is carried out assuming that the optimal model will show optimal performance (e.g., 100\%). However, in the game snapshot data problem, it can be seen that even the optimal model cannot achieve 100\% accuracy due to the uncertainty of the game snapshot data itself as shown in Fig.~\ref{figure1}.
Compare to the evaluation of a classification problem that can be measured with a given ground-truth label, proper evaluation of the probability estimation model is more challenging because the ground-truth probability is unknown.

Recently in fields other than esports, two representative works namely, the Brier score~\cite{brier1950verification} and the Expected Calibration Error ($ECE$)~\cite{naeini2015obtaining} are proposed to measure the calibration performance of models instead of accuracy. 
Calibration refers to the statistical consistency between the estimated probabilities and the true results. 
Unlike accuracy, measuring the calibration performance thus can reflect the estimated probability values.

In this section, we firstly formulate win probability estimation problem in the esports field. Subsequently, we introduce two representative metrics for calibration performance measures based on their definitions, advantages, and disadvantages. Finally, we propose a novel metric called Balance score motivated by two metrics.

\subsection{Problem Formulation}
Assume the game snapshot dataset consists of feature vectors
$\bm{x_{i}}, 1 \le i \le n$, and their corresponding results $y_{i} \in \mbox{ }\{0, 1\}$. 
For $n$ game snapshots, each feature vector can consist of several scalar information at the time (e.g., earned gold, earned experience points) depending on the game to measure. 
Following, the `0' and `1' for the result $y_{i}$ respectively refers to the `lose' and `win' at the end of the game.

Given a win probability estimation model, the model predicts the win probability $\hat{p_{i}}$ of each snapshot $\bm{x_{i}}$. Noting that the real win probability $p_{i}$ of the snapshot is unknown. 
The purpose of the evaluation metric for the win probability estimation model is thus to adequately evaluate the model's predicted output $\hat{p}$ of each game snapshot.

\subsection{Brier Score}
Scoring function is a set of rules that involves computation between the estimated probability and the actual outcome. 
Scoring function can be used to evaluate the estimated probabilities and encourage models to estimate `good' probabilities by providing the appropriate score~\cite{winkler1996scoring}. 
For each snapshot \begin{math}\bm{x_{i}}\end{math}, only \begin{math}\hat{p_{i}}\end{math} and \begin{math}y_{i}\end{math} are used as input of scoring function since \begin{math}p_{i}\end{math} is unknown. 
The Brier score is one of the representative scoring rules which has been normally used to evaluate the probability estimation models. 
The Brier score can be represented with its scoring function $f_{br}(\cdot)$ and expectation as follows: 
\begin{equation}
Brier\mbox{ }score = \frac{1}{n}\sum_{i=1}^{n} f_{br}(\hat{p_{i}}, y_{i}),
\label{eq1}
\end{equation} where
\begin{equation}
f_{br}(\hat{p_{i}}, y_{i})=(\hat{p_{i}}-y_{i})^2.
\label{eq2}
\end{equation}
Equation~\eqref{eq1} can be decomposed again into two terms as follows~\cite{rufibach2010use}: 
\begin{equation}
\begin{split}
Brier\mbox{ }score = \frac{1}{n}\sum_{i=1}^{n} (\hat{p_{i}}-y_{i})(2\hat{p_{i}}-1) + \frac{1}{n}\sum_{i=1}^{n} \hat{p_{i}}(1-\hat{p_{i}}).
\end{split}
\label{eq3}
\end{equation} 
In equation~\eqref{eq3}, the left term falls into 0 in expectation under the perfect calibration while the right term related to the sharpness, the concentration of the predictive distribution~\cite{gneiting2007strictly}. This means that the Brier score simultaneously addresses the calibration performance and the sharpness of probability estimation. 
Also, the Brier score is one of the strictly proper scoring rules~\cite{schervish1989general} which have the characteristic that estimating $\hat{p_{i}}$ to $p_{i}$ is the only optimal strategy for the expected score. 
In general, the expected score of a model under a certain scoring function can be calculated as: 
\begin{equation}
\begin{split}
  {\mathbb E_{model}}[score]
  &= \int_{0}^{1} \int_{0}^{1} 
      \left[ p \cdot f(\hat{p}, 1) + (1-p) \cdot f(\hat{p}, 0) \right] \\
  &\quad \times {P_{model}}(\hat{p}|p) d\hat{p} \, \pi(p)dp
\end{split}
\label{eq4}
\end{equation}
where $f(\hat{p}, y)$, $\pi(p)$ and $P_{model}(\hat{p}|p)$ respectively denotes the scoring function, the distribution of the ground-truth $p$ on target dataset, and the conditional probability of \begin{math}\hat{p}\end{math} under $p$ of the model.

An optimal model which always gets $\hat{p_{i}}=p_{i}$ can also be evaluated in terms of the Brier score. Assume a situation where the optimal model gets scores with uniform distribution of the ground-truth \begin{math}p\end{math}(i.e., \begin{math}\pi(p)=1\end{math} for 0 $\leq p \leq 1$, $\pi(p)=0$ otherwise). 
In the case of esports, it can be seen as an example of win probability estimation in the middle of a match where the win probability is likely to be uniformly distributed rather than concentrated. 
In such conditions, optimal expected scores of accuracy (which also can be represented as a scoring function) and the Brier score can be calculated as 0.75 (75\%) and 0.166 respectively, according to the equation~\eqref{eq4}. Since the operating condition of a specific dataset is unknown in real cases, the optimal score which can be the target is also unknown. It means that the low Brier score reported in previous studies cannot guarantee general performance. 
Instead, the score can only be used as a relative measure of multiple models for a single fixed dataset that shares an operating condition. 
This limitation of the Brier score can be a major drawback to its adoption as a metric in the field of esports which does not have a fixed dataset and the operating condition varies widely each time.

\subsection{Expected Calibration Error}
In recent studies, several approaches such as the reliability diagram~\cite{degroot1983comparison}, Expected Calibration Error ($ECE$), and Maximum Calibration Error ($MCE$)~\cite{naeini2015obtaining} have been proposed to measure the calibration performance of a model. Among these metrics, $ECE$ is frequently used because it can reasonably express the calibration performance of the model with a single scalar value. The perfect calibration from the model can be expressed as follows: 
\begin{equation}
    Prob(\hat{Y} = 1|\hat{P} = p) = p, \forall p \in [0,1].
\end{equation} Then, the model's true expected calibration error is represented as follows: 
\begin{equation}
    \mbox{True ECE} = \underset{\hat{P}}{\mathbb{E}} [|Prob(\hat{Y}=1|\hat{P}=p)-p|].
    \label{eq6}
\end{equation}
To approximate the true expected calibration error, Guo et al.~\cite{guo2017calibration} suggested dividing the set of \begin{math}\hat{p_{i}}\end{math} with [0, 1] probability interval into $M$ equally spaced bins. 
$ECE$ value is calculated based on the errors from these bins as follows: \begin{equation}
    ECE = \sum_{m=1}^{M} \frac{|B_{m}|}{n} |\overline{y}(B_{m}) - \overline{\hat{p}}(B_{m})|,
    \label{eq7}
\end{equation} 
where \begin{math}B_{m}\end{math} denotes the set of indices of predictions belonging to m-th bin, \begin{math}\overline{y}(B_{m})\end{math} denotes the proportion of the true results of predictions in m-th bin, and \begin{math}\overline{\hat{p}}(B_{m})\end{math} denotes the average of probability predictions in m-th bin. Noting here the term `$ECE$' refers to the approximation of the true expected calibration error (True ECE) by equation~\eqref{eq7} hereafter. Compare to accuracy and the Brier score, an optimal model which always predicts \begin{math}p_{i}\end{math} as \begin{math}p_{i}\end{math} can get 0 $ECE$ value. Knowing the optimal value has the advantage that the model experimenter can check whether the model approaches perfect calibration by tracking the $ECE$ value of the model~\cite{guo2017calibration}. 

However, $ECE$ has some limitations. First of all, there is no criterion for determining how many bins to divide. To get a precise approximation of the true expected calibration error, a larger $M$ would be better. However, if $M$ is increased, the number of \begin{math}\hat{p_{i}}\end{math} in each bin decreases which results in a large bias in the $ECE$ value~\cite{nixon2019measuring}. If the calibration performance of several models on one dataset is compared, the order of the calibration performance of the models can be changed as the $M$ value is changed.
Also, due to the nature of the calibration metric which divides bins and collects predictions to calculate values, only evaluations on the entire dataset are possible. Based on these observations, we propose a scoring function based metric called Balance score which addresses the shortcomings of existing metrics and takes only their advantages.

\subsection{Balance Score}
The Balance score is a score with gain and loss strategy, which pursues the balance of the score. 
If model predicts the true observation \begin{math}y_{i}\end{math} based on \begin{math}\hat{p_{i}}\end{math}, it gains a score and if it predicts incorrectly, it loses a score. 
For each \begin{math}\bm{x_{i}}\end{math} which is difficult to predict the result correctly (e.g., $p$ in 40\% $\sim$ 60\%), a large score is gained if the predicted result is correct, and a small score is lost if the predicted result is incorrect. 
Conversely, when \begin{math}\bm{x_{i}}\end{math} is easy to predict the result correctly (e.g., $p$ close to 0\% or 100\%), a small score is gained if the predicted result is correct, and a large score is lost if the predicted result is incorrect. 
The Balance score with its scoring function $f_{ba}(\cdot)$ can be defined as follows:
\begin{equation}
f_{ba}(\hat{p_{i}}, y_{i})=
\begin{cases}
1-\hat{p_{i}}, & \mbox{if }\hat{p_{i}} \ge 0.5 \mbox{ and } y_{i} = 1 \\
\hat{p_{i}}, & \mbox{if }\hat{p_{i}} < 0.5 \mbox{ and } y_{i} = 0 \\
-\hat{p_{i}}, & \mbox{if }\hat{p_{i}} \ge 0.5 \mbox{ and } y_{i} = 0 \\
-1 + \hat{p_{i}}, & \mbox{if }\hat{p_{i}} < 0.5 \mbox{ and } y_{i} = 1
\end{cases},
\end{equation}
\begin{equation}
Balance\mbox{ }score = 
\frac{1}{n}\sum_{i=1}^{n} f_{ba}(\hat{p_{i}}, y_{i}).
\end{equation} 
To give new properties of the Balance score, let \begin{math}G(p)\end{math} be the pointwise expected score when model predicts \begin{math}p\end{math} as \begin{math}p\end{math}. Then the model can get 0 score by maintaining the total score to be balanced as follows:
\begin{equation}
\begin{split}
    G(p) &= pf_{ba}(p,1) + (1-p)f_{ba}(p,0)\\
      &= 0  \text{ for }  \forall p \in [0, 1] .
\end{split}
\label{eq10}
\end{equation} 
Also, when \begin{math}p\end{math} is estimated to be a different value, the balance is broken in proportion to the difference between \begin{math}p\end{math} and the estimated value. More generally, let \begin{math}g(q;p)\end{math} be the pointwise expected score function when model predicts \begin{math}q\end{math} under the ground-truth probability \begin{math}p\end{math}. Then \begin{math}G(p) = g(p;p)\end{math} holds. Also, the following expression holds:
\begin{equation}
    |g(q;p)| = |q-p| \text{ for }  \forall q,p \in [0, 1].
\label{eq11}
\end{equation}

Equation~\eqref{eq11} simply shows that model gets 0 score only if model estimates \begin{math}p\end{math} under \begin{math}p\end{math}, and the balance is broken in proportion to the difference. The Balance score is not following the proper scoring rule suggested in \cite{schervish1989general} because it is a new scoring rule with gain and loss. However, the Balance score still shares the concept of a proper scoring rule that estimating \begin{math}p\end{math} as \begin{math}p\end{math} is the optimal strategy for the expected score. 

An expected score of the Balance score can also be calculated with equation~\eqref{eq4}. Same with the $ECE$, the optimal model also can get 0 value according to equation~\eqref{eq10} regardless of the operating condition. This means that the optimal Balance score can be the target of models to be trained. 
Moreover, recent machine learning models suffer mainly from overconfidence or underconfidence in terms of probability estimation~\cite{guo2017calibration}. According to the tendency of the model, if a extreme prediction \begin{math}\hat{p_{i}}\end{math} close to 0 or 1 is given compared to the actual \begin{math}p_{i}\end{math}, it is called overconfident, and if it is given a mild value, it is called underconfident. Assume a situation with \begin{math}
Prob(\hat{Y}=1|\hat{P}=p) = q_{p}\end{math}. If a model has an overconfident property, $p$ is placed in \begin{math}0.5\le p\le q_{p} \mbox{ for } 0.5 \le p \end{math} and \begin{math}  q_{p}\le p < 0.5 \mbox{ for } p< 0.5\end{math}. 
Then true expected calibration error is approximated as: \begin{eqnarray*}
\mbox{True ECE} &=&  \underset{\hat{P}}{\mathbb{E}} [|Prob(\hat{Y}=1|\hat{P}=p)-p|]\\
&=& \underset{\hat{P}}{\mathbb{E}}[|q_{p}-p|]\\
&=& \underset{\hat{P}<0.5}{\mathbb{E}}[p-q_{p}] + \underset{0.5\le\hat{P}}{\mathbb{E}}[q_{p}-p]\\
&=& \underset{\hat{P}<0.5}{\mathbb{E}}[-g(q_{p};p)] + \underset{0.5\le\hat{P}}{\mathbb{E}}[-g(q_{p};p)]\\
&=& -\underset{\hat{P}}{\mathbb{E}}[g(q_{p};p)]\\
&\approx& -Balance \mbox{ } score.
\end{eqnarray*} Conversely, if model has an underconfident property, \begin{math}\mbox{True ECE} \approx Balance \mbox{ } score\end{math}.
This means that the Balance score is another approximation of true expected calibration error under general condition, but without binning technique.
To conclude, Table~\ref{table1} summarizes whether the metrics have good properties as probability estimation metric.

\begin{table*}[t]
\centering
\caption{A summary of the candidate metrics for evaluating probability estimation models.}
\label{table1}
\begin{tabular}{c|cccc}
\noalign{\smallskip}\noalign{\smallskip}\hline\hline
\multicolumn{1}{c|}{Properties}                         & \multicolumn{1}{l}{Accuracy} & \multicolumn{1}{l}{Brier score} & \multicolumn{1}{l}{\textit{ECE}} & \multicolumn{1}{l}{Balance score} \\ \hline
Correctly estimating $p$ is its optimal strategy & \Large{$\circ$}       & \Large{$\circ$}                                         & \Large{$\circ$}                                     & \Large{$\circ$}                                         \\ \hline
Can evaluate a single estimation               & \Large{$\circ$}                             & \Large{$\circ$}                                         & $\times$                                          & \Large{$\circ$}                                           \\ \hline
The optimal value is given                     & $\times$                             & $\times$                                         & \Large{$\circ$}                                          & \Large{$\circ$}                                           \\ \hline
Can be an absolute measure                     & $\times$                             & $\times$                                         & \Large{$\circ$}                                          & \Large{$\circ$}                                           \\ \hline
Do not have hyperparameter                  & \Large{$\circ$}                             & \Large{$\circ$}                                         & $\times$                                          & \Large{$\circ$}                                           \\ \hline
Is a calibration measure                            & $\times$                             & \Large{$\circ$}                                         & \Large{$\circ$}                                          & \Large{$\circ$}                                           \\ \hline\hline
\end{tabular}
\end{table*}

\section{Empirical Results}
In this section, we evaluate the proposed evaluation metric using simulation studies and real game snapshot data. 
Under the first case study, we evaluate the expected accuracy, Brier score, $ECE$, and Balance score obtained from the optimal model under various beta distributions. The expected scores obtained from the logistic regression model under real game snapshot datasets at different time points are also evaluated to show the limitations of accuracy and the Brier score. In the second case study, we compare two calibration based metrics ($ECE$, Balance score) in detail. 

\subsection{Case Study I: Limitations of accuracy and Brier score}\label{tt}
In equation~\eqref{eq4}, an analytical method to calculate the expected score of a model under a specific operating condition and a specific scoring function was presented. Assume a situation in which a model was scored on a dataset following a specific distribution. This is done by repeating the situation where $p_{*}$ is firstly generated following a specific distribution, $y_{*}$ is determined according to the value of $p_{*}$, and assume the model estimates it to $\hat{p_{*}}$. Since each $p_{*}$ exists between [0,1] and the conditions of the two competing teams are the same, the win probability distribution of the game snapshot dataset can be considered to follow a symmetric beta distribution with an average win probability of 0.5 (50\%). The win probability distribution of the game state snapshot dataset is very different depending on the state of the game. At the beginning of the game, the odds of the two competing teams are not significantly different, so the win probabilities will be concentrated near 50\%. As time goes by, the tail of the distribution will become thicker, and the distribution of the win probability will lean to both extreme sides after the middle of the game. This situation can be simulated by adjusting the $\alpha$ and the $\beta$ parameters of the beta distribution. 

As a simulation study, The upper row of the Table~\ref{table2} summarizes expected accuracy, Brier score, $ECE$, and Balance score of the optimal model on dataset with the distribution of probabilities following $Beta(0.5, 0.5)$, $Beta(1, 1)$, and $Beta(2, 2)$. By generating 100,000 synthetic data $p_{*}$ following each distribution, the score can be calculated according to the simulation method which becomes the approximation of the equation~\eqref{eq4}. The number of bin $M$ for $ECE$ is set to 10 and subsequently calculated by collecting pairs of ($\hat{p_{*}}$, $y_{*}$).


As shown in the upper row of the Table~\ref{table2}, the optimal model cannot achieve 100\% accuracy, and the optimal accuracy value varies greatly depending on the distribution, from 68.92\% to 81.88\%. Similar to the accuracy, the Brier score also differs greatly depending on the distribution (0.1995 to 0.1241). Also, the true $p$ and the true distribution of $p$ of actual dataset are unknown as mentioned in the previous section. Therefore, the result of achieving a certain accuracy and Brier score in a certain dataset cannot be a general performance. Since the true ECE of the optimal model is 0, both the $ECE$ and the Balance score properly approximate them to 0 regardless of the distribution of $p_{*}$. Noting that the error of $ECE$ is slightly larger than the Balance score due to the bias caused by the binning process.

\begin{figure}[ht]
\centering
    \includegraphics[width=0.22\textwidth]{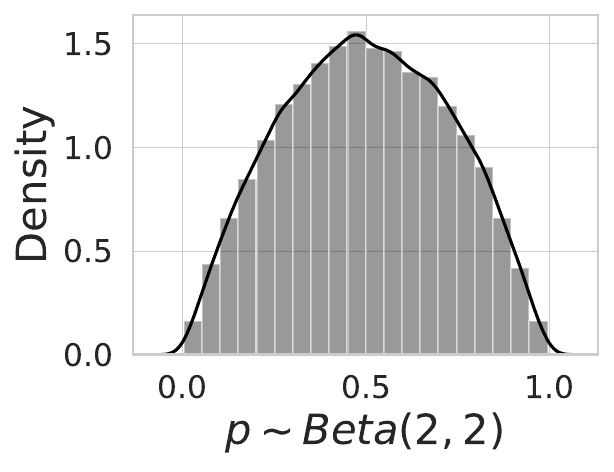}%
    \hfill
    \includegraphics[width=0.22\textwidth]{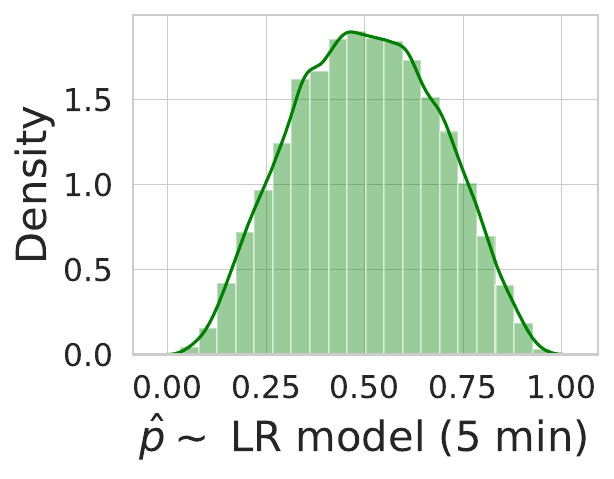}%
    \hfill
    \includegraphics[width=0.22\textwidth]{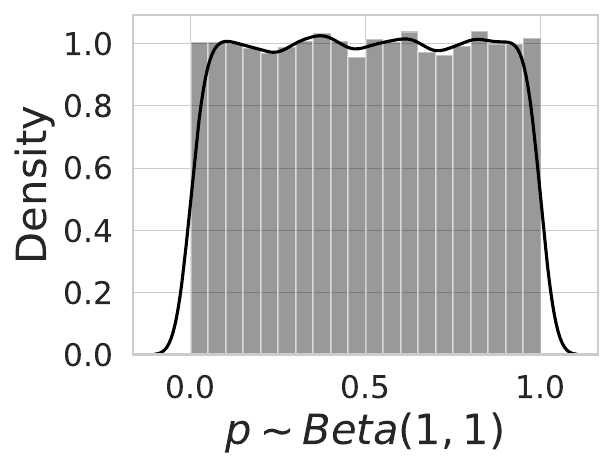}%
    \hfill
    \includegraphics[width=0.22\textwidth]{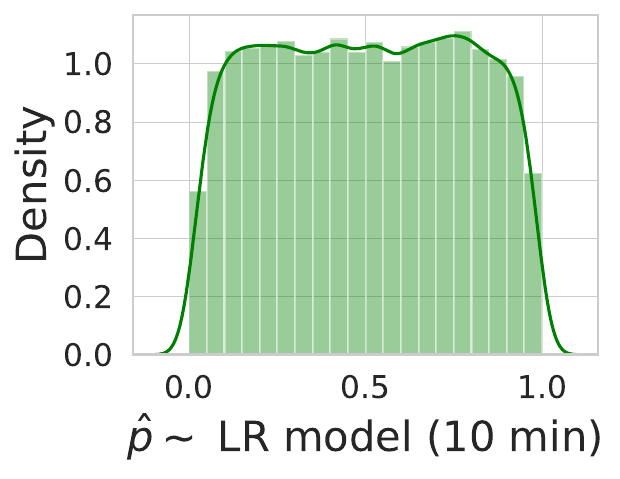}%
    \hfill
    \includegraphics[width=0.22\textwidth]{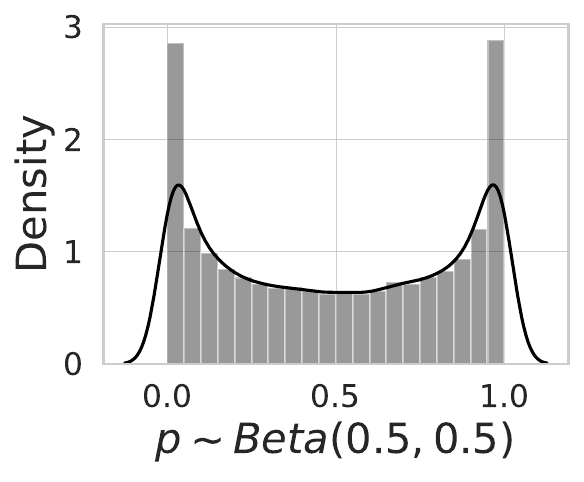}%
    \hfill
    \includegraphics[width=0.22\textwidth]{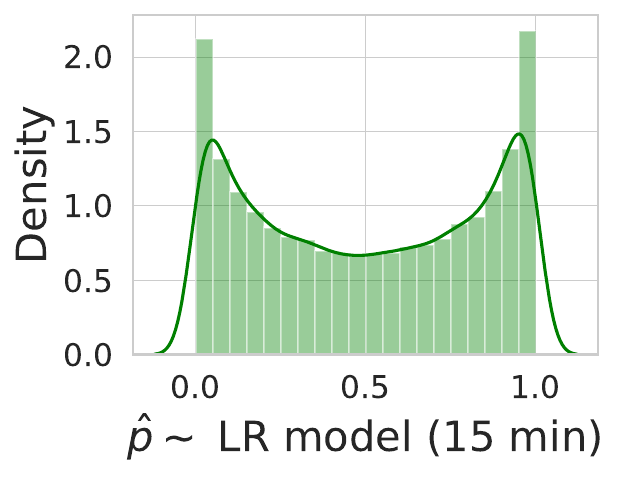}%
    \caption{Three plots in left side respectively refers to the plot of $p$ obtained from the generated beta distributions $Beta(2, 2)$, $Beta(1, 1)$, and $Beta(0.5, 0.5)$. Distributions of $\hat{p}$ derived by the logistic regression model trained with the real game snapshot datasets at 5, 10, and 15 minutes are also plotted on the right side.} 
	\label{figure3}
\end{figure}

For the experiment of real game snapshot data, the game snapshot data from the ``League of Legends'', one of the popular MOBA games is collected through the Riot Games public API \cite{riotapi} and evaluated. We collected datasets at three different time points which are expected to follow the three distributions assumed in the simulation study. 
The feature vector $x_*$ for each snapshot data is generated by taking 14 indicators that affect the winning of the match (the gold difference for each role player (5), the experience difference for each role player (5), the number of killed dragons for each team (2), and the number of destroyed towers for each team (2) similar to that in \cite{maymin2021smart}. Following our problem formulation, their corresponding results $y_*$ are also recorded as the true labels. For each time point, 100,000 matches were taken and vectorized, while 60,000 matches were allocated as a training set and the remaining 40,000 matches were allocated as a testset. The lower row of the Table~\ref{table2} shows the results calculated by each metric after the learning with a logistic regression model that naturally derives probabilities.
The distribution of $\hat{p}$ derived by the logistic regression model trained with the dataset for each time period is shown in Fig.~\ref{figure3}. As shown in the figure, the game snapshots at 5, 10, and 15 minutes respectively resemble the distributions of $Beta(2, 2)$, $Beta(1, 1)$, and $Beta(0.5, 0.5)$. 


\begin{table}
\centering
\caption{Expected scores obtained from the optimal model (under beta distributions) and the logistic regression model (for each time point).}
\label{table2}
\begin{tabular}{c|cccc}
\noalign{\smallskip}\noalign{\smallskip}\hline\hline
Distributions \& Time points & Accuracy & Brier & $ECE$ & Balance\\
\hline
$Beta(0.5, 0.5)$ & 81.88\%  & 0.1241 & 0.0019 & 0.0000 \\
$Beta(1, 1)$ & 75.01\%   & 0.1666 & 0.0017 & -0.0004\\
$Beta(2, 2)$ & 68.92\%   & 0.1995 & 0.0031 & 0.0013\\
\hline
\hline
15 min               & 79.84\%    & 0.1385 & 0.0078 & -0.0039 \\
10 min               & 73.55\%   & 0.1755 & 0.0068 & -0.0016\\
5 min                & 65.56\%   & 0.2159 & 0.0058 & 0.0043\\
\hline
\hline
\end{tabular}
\end{table}

Consider the distributions of $\hat{p}$ from the trained model are similar to the assumed distributions of $p$ and the scores from the trained model are similar to the optimal model’s score. Based on the results in the Table~\ref{table2}, the trained logistic regression model can be considered a pretty nice model. However, the low accuracy of 65.56\% can be considered that the model's discrimination power is insufficient. In a similar manner, the high Brier score of 0.2159 may seem insufficient to use the logistic regression model as a probability estimation model. For this gap between the understandings of the model's performance, it would be right to understand that the performance is limited by the uncertainty of the data rather than conclude that the logistic regression model does not have sufficient capacity or that there is a learning problem. However, since the distribution of true $p$ is unknown and subjective, comparing the performance with that of the optimal score on the assumed distribution is also not reasonable. Instead, the $ECE$ or the Balance score has a chance to be an absolute measure of the model's performance which also measures the calibration performance of the model. Since we know that the optimal value of both metrics is 0, we can directly understand the performance of the logistic regression model itself and thus can target to update the model targeting the values to be 0.


\subsection{Case Study II: Comparison between $ECE$ and Balance score}

$ECE$, one of the best choices for approximating the true ECE, involves a binning process (see equation~\eqref{eq7}).
The binning hyperparameter $M$ should be increased to make the approximation more accurate, but the bias occurs because the number of $\hat{p_{*}}$ in each bin decreases.

\begin{figure}[ht]
\centering
	\includegraphics[width=1.0\columnwidth]{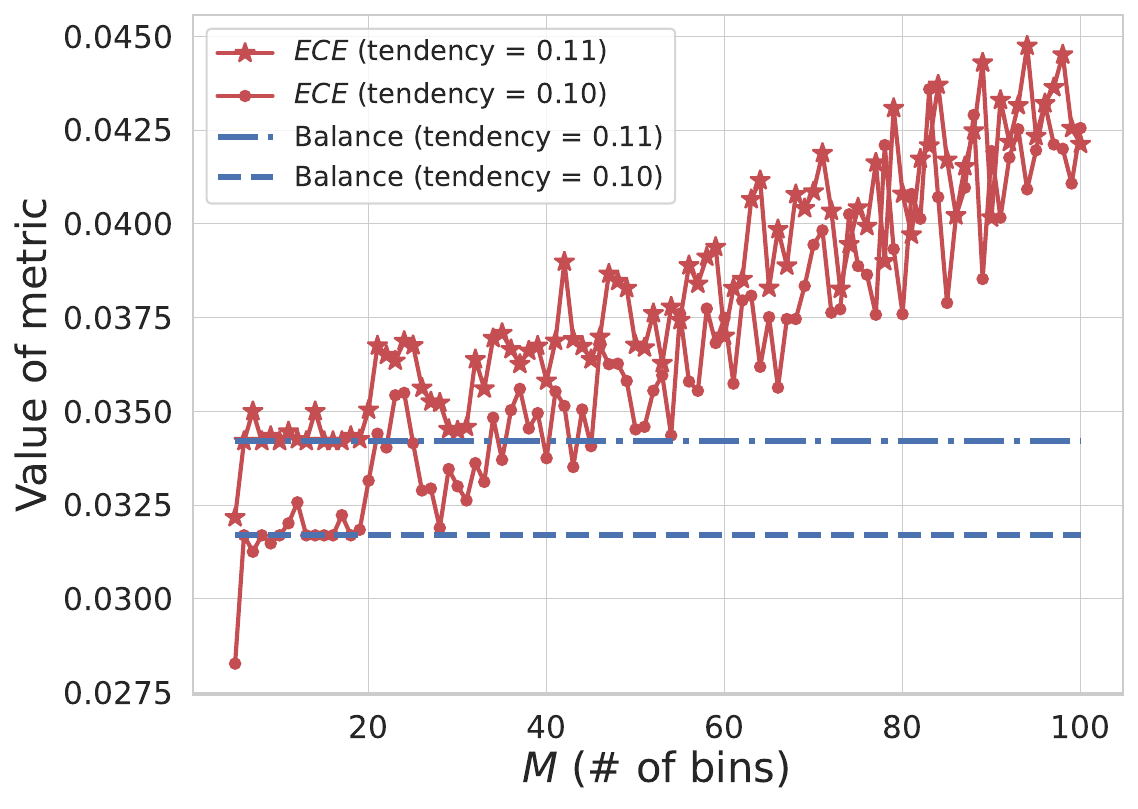}
	\caption{$ECE$ values with increasing binning number $M$ and the absolute Balance score of two overconfident models. Red lines with circle markers and asterisk markers respectively refers to the $ECE$ of the model with 0.1 and 0.11 tendency. Blue dashed line and dash-dotted line respectively refers to the absolute Balance score of the model with 0.1 and 0.11 tendency.}
	\label{figure5}
\end{figure}

To illustrate limitations of a binning process for performance evaluation task, we assume two synthetic models with some degree of overconfident tendency.
These models tend to overestimate for each true $p$. 
When the tendency is 0.1, the degree is about 1/10 of the model that is completely overconfident and estimates 0 or 1 for all $p$. 
For example, the model estimates $0.9*0.6 + 0.1*1 = 0.64$ for true $p$ of 0.6, and $0.9*0.2 + 0.1*0 = 0.18$ for true $p$ of 0.2. 
Fig.~\ref{figure5} shows the resultant $ECE$ values along the increasing $M$ from 5 to 100 with 10,000 ($\hat{p_{*}}$, $y_{*}$) pairs obtained from the synthetic models with a tendency of 0.1 and 0.11. The corresponding 10,000 true $p_{*}$ are generated from the uniform distribution. Since the true ECE of the model with tendency 0.1 is smaller than the model with tendency 0.11, it is clear that the $ECE$ value also should be smaller. However, depending on the selection of $M$, $ECE$ of the model with a tendency of 0.11 can be smaller than the model with a tendency of 0.1 as shown in the crossing points of blue and red lines in Fig.~\ref{figure5}. 
This shows that the order of the models' performance can be changed by the subjective choice of the experimenter regardless of their actual performance. Instead, the Balance score can be an exact measure for the performance evaluation of the model without any hyperparameter.

\begin{figure}[ht]
	\includegraphics[width=1.0\columnwidth]{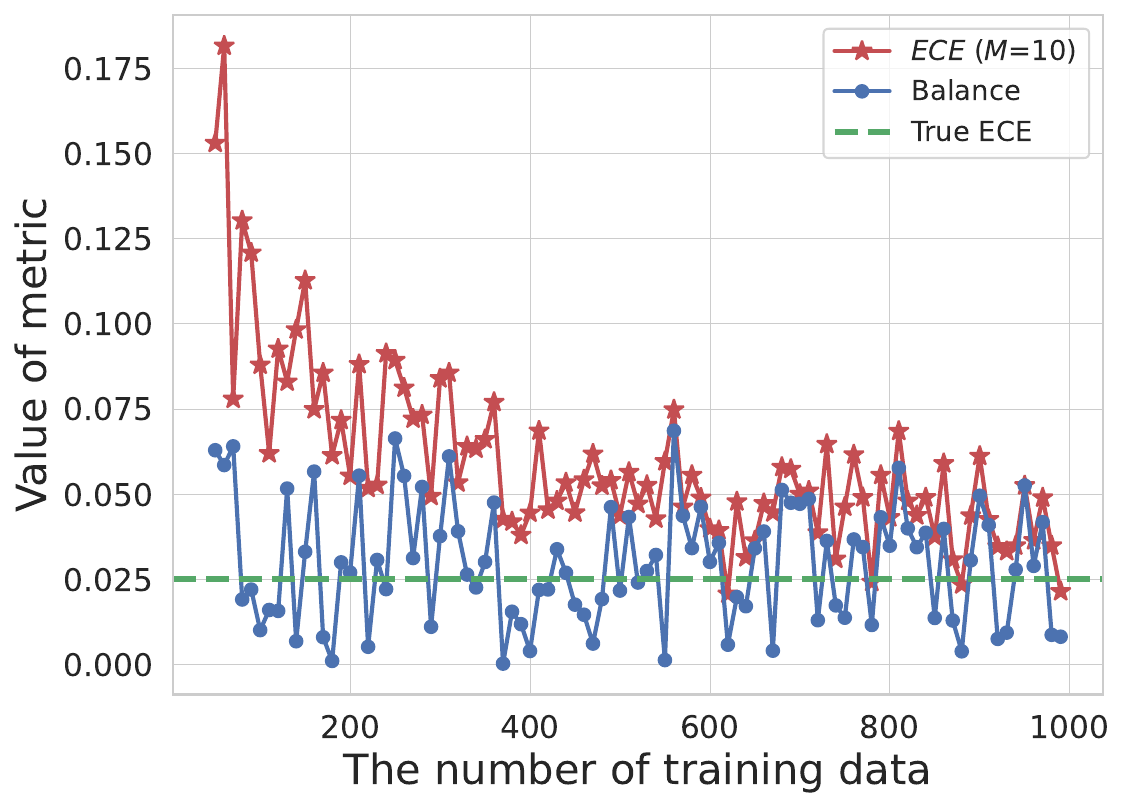}
	\caption{$ECE$ and the absolute Balance score of overconfident model with 0.1 tendency over the increasing data size. The red line with asterisk markers and the blue line with circle markers respectively refers to $ECE$ and Balance score while the green dashed line denotes the analytic true ECE value of the evaluated model.} 
	\label{figure6}
\end{figure}

In addition, the Balance score also requires much fewer data to estimate the true ECE based on pointwise calculation.
When the distribution of $p_{*}$ is uniform, the analytic solution of the true ECE for an overconfident model with a tendency of 0.1 can be calculated to 0.025 following the equation~\eqref{eq6}. 
Fig.~\ref{figure6} shows the $ECE$ ($M=10$) and the Balance score of the model along the increment of utilized synthetic data size from 50 to 1000. Compare to the $ECE$ which needs more than 500 sample data to approximate the true ECE, the Balance score can approach the true ECE value with much fewer data. Note that the pointwise calculation of the Balance score also resulted in the computational efficiency compared to the $ECE$ calculation. Based on the observations, the Balance score without a subjective binning process shows several advantages over the $ECE$ in terms of evaluation metric.

\section{Conclusion}
In this work, we have investigated how to adequately evaluate probability estimation models via esports' win probability estimation model.
Through the theoretical analysis and experiments, we found that a novel metric called Balance score, motivated by the Brier score and $ECE$, takes the advantages of existing metrics and also solves their shortcomings.
Also, under machine learning models’ general condition, we found that the Balance score can be an effective approximation of the true expected calibration error.

In future works, we expect to develop a model that provides more reliable probabilities for esports' win probability with the help of proper evaluation by the Balance score.
Additionally, we will investigate favorable effects of replacing $ECE$ in various calibration-involved areas such as calibrated model learning and post-processing calibration methods.

\bibliographystyle{IEEEtran}
\bibliography{pakdd2023}

\end{document}